\documentclass[11pt]{article}

\usepackage[]{acl}

\usepackage{times}
\usepackage{latexsym}
\usepackage{amsmath}
\usepackage{booktabs}
\usepackage{multirow}
\usepackage{colortbl}
\usepackage[table]{xcolor}
\usepackage{amsfonts}
\usepackage{enumitem}

\usepackage[T1]{fontenc}

\usepackage[utf8]{inputenc}

\usepackage{microtype}

\usepackage{inconsolata}

\usepackage{graphicx}

\title{Circuit Distillation}

\author{\textbf{Somin Wadhwa}\quad\quad \textbf{Silvio Amir}\quad\quad \textbf{Byron C. Wallace} \\ 
Northeastern University  \\ 
\texttt{\{wadhwa.s, s.amir, b.wallace\}@northeastern.edu} 
}

\begin{document}
\maketitle
\begin{abstract}
Model distillation typically focuses on \emph{behavioral} mimicry; a student model is trained to replicate a teacher's output while treating its internal computations as a black box. 
We propose an alternative approach: Distilling the underlying computational mechanisms implemented by a teacher model. 
Specifically, we introduce \emph{circuit distillation}. 
This works by first aligning ``functionally correspondent'' circuit components in teacher and student models, and then 
optimizing an objective that encourages these aligned components to induce similar internal representations.  
We evaluate circuit distillation on entity tracking and theory of mind (ToM) tasks using {\tt Llama3} models. 
Our results demonstrate that circuit distillation outperforms standard  distillation, successfully transferring algorithmic capabilities by adjusting only a small, targeted subset of student model parameters. 
This work establishes the feasibility of transferring mechanisms, which may in turn enable efficient distillation of targeted teacher capabilities via interpretable mechanism transfer. 
\end{abstract}

\section{Introduction}

\emph{Model distillation} entails 
training a relatively small, efficient ``student'' LM using a larger and more capable teacher LLM \citep{gou2021knowledge}. 
The prevailing training paradigm is behavioral mimicry: The student model is optimized to replicate the output of the large ``teacher'' model (usually minimizing divergence between final layer logits; \citealt{hinton2015distillingknowledgeneuralnetwork}). 
More recent work has has focussed on distilling ``reasoning'' capabilities \citep{shridhar2023distilling,li2023symbolic,wadhwa-etal-2024-investigating}.

Distillation permits effective transfer of task-specific knowledge \citep{xu2024survey}. But the standard transfer mechanism is fundamentally bottlenecked: the student learns only from teacher \emph{outputs}, and must infer for itself how to perform the task. A more direct approach is to let the student learn from the algorithms the teacher implements internally. This is the idea we explore here.

More specifically, work in \emph{mechanistic interpretability} \citep{saphra2024mechanistic} has established that transformer models sometimes implement algorithms via subgraphs of attention heads and MLP layers, i.e., ``circuits'' \citep{shi2024hypothesis}. 
In light of this, we propose to guide student models to functionally emulate relevant teacher circuits.  
We call this \emph{circuit distillation}. 
Our hypothesis is that enforcing alignment at the component level enables distillation of not only the teacher’s knowledge, but also its internal algorithms.

\begin{figure*}
\centering
\includegraphics[scale=0.425]{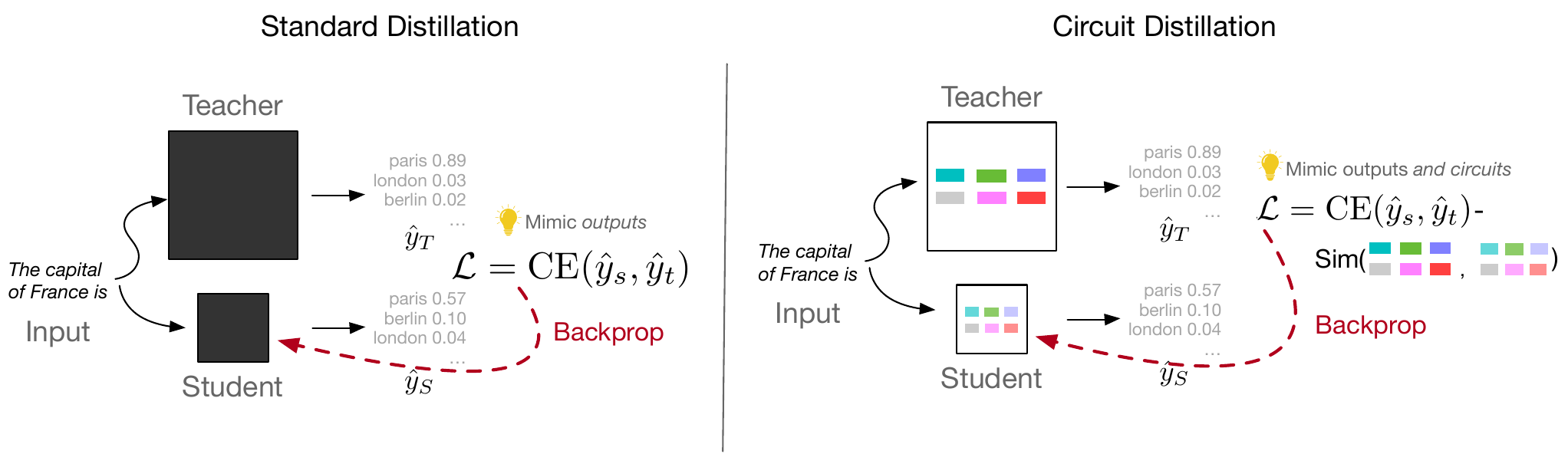}
\caption{We propose \emph{circuit distillation} which entails training a small student model to mimic not only the outputs of a larger teacher model (left is the standard distillation setting; \citealt{hinton2015distillingknowledgeneuralnetwork}), but also its internal circuitry, i.e., \emph{how} it executes a task of interest (right).}
\label{fig:main_illustration}
\end{figure*}

Realizing circuit distillation poses two main technical challenges: (1) establishing a functional correspondence between components in models of different sizes, and, (2) designing an objective that enforces representational alignment between these circuits. 
We develop methods for both, and report promising empirical results. 
We view circuit distillation as an interesting direction, but acknowledge upfront that the paradigm is only applicable in settings where:  
(i) the teacher is white-box, (ii) there exists a known circuit that plausibly plays a role in the target capability we hope to transfer to the student, and (iii) the student shows a weak version of that capability, so that we can align components. 
The goal is to transfer a specific, understood mechanism efficiently.\footnote{Distilling very broad, distributed capabilities such as general next-token prediction---where the relevant ``circuit'' may span most of the model---would be contrary to the aim of transferring legible small modules.} 
As more circuits are identified in strong open-weight models, this approach may become increasingly practical. 

We offer the following contributions.
(1) We introduce \emph{circuit distillation} (Figure \ref{fig:main_illustration}), a new form of mechanistic distillation that modifies the student objective to promote alignment of internal circuits, not just output-level mimicry.
(2) We propose a functional component-mapping strategy using \emph{ablation impact similarity} to address the circuit correspondence problem, providing an intuitive method for identifying analogous attention heads across differently scaled models.
(3) We introduce a transformation-invariant loss term that enforces representational similarity between mapped circuit-head representations during training.
(4) We validate the approach on two tasks: entity tracking \citep{prakash2024finetuningenhancesexistingmechanisms} and causal Theory of Mind (ToM; \citealt{prakash2025languagemodelsuselookbacks}). 
We find that circuit distillation can transfer mechanisms, enabling student models to perform tasks using internal mechanisms aligned with their teachers, and outperforming models distilled from teacher outputs alone.

Our findings indicate that it is feasible to distill not just knowledge, but also the algorithms that produce it. By shifting the focus of distillation from outputs to circuits, we take a concrete step toward building student models whose internal computations we can better understand and direct. 

\section{Methods}
\label{sec:methods}
In this section we describe our approach to model distillation from a mechanistic perspective. Our goal is for the student to emulate not only the teacher’s outputs, but also its internal mechanisms, and in particular relevant \emph{circuits} \citep{conmy2023towards,shi2024hypothesis}. To do this, we first define a way to quantify representational similarity at the level of specific circuit heads (Section \ref{sec:rep-sim}), and then incorporate this measure into a distillation objective (Section \ref{sec:distill-obj}) that is combined with the standard distillation loss (Section \ref{sec:loss}).

\subsection{Quantifying Representational Similarity}
\label{sec:rep-sim}

Understanding and transferring learned algorithms from a teacher to a student model requires a method to compare their internal representations. 
We focus on individual attention heads or MLP sub-layers that have been identified as participating in a circuit in the teacher model. 
Metrics like cosine similarity between different (student and teacher) networks are not inherently meaningful given that dimensions may differ and are anyways arbitrary (e.g., due to basis rotations or isotropic scaling). 

Therefore, we adopt Centered Kernel Alignment (CKA; \citealt{kornblith2019similarity}), which provides a robust and theoretically grounded measure of representational similarity between model internals.  
CKA is invariant to orthogonal transformations (including permutations) and isotropic scaling of the feature space, making it well-suited to compare activations from different architectures or---relevant to our work---specific corresponding circuit heads in teacher and student  of different sizes. 
This invariance allows us to compare the \textit{functional similarity} of these circuit components. 
The intuition behind CKA is to measure the 
similarity between representational similarities induced by networks: Networks are ``similar'' if they produce representations that yield comparable pairwise similarities between examples \citep{kornblith2019similarity}.

More precisely, given a batch of $m$ inputs, let $X \in \mathbb{R}^{m \times p_1}$ be the activation matrix from a specific circuit head in the student model, and $Y \in \mathbb{R}^{m \times p_2}$ be the activation matrix from the corresponding (or targeted) circuit head in the teacher model. 
Here $p_1$ and $p_2$ are the dimensions of the activations for the respective circuit heads.
Concretely, a head's activation is its output before the attention output projection mixes heads (i.e., its slice of the input to $W_O$), taken at the final token of each input.
We compute Gram matrices $K = XX^\top$ and $L = YY^\top$
and then measure correlations between the pairwise similarities induced by the two networks. 
Specifically, we adopt the Hilbert-Schmidt Independence Criterion (HSIC), a kernel-based measure of independence between variables\footnote{We use a linear kernel.}. 
Given two kernel matrices $K, L \in \mathbb{R}^{m \times m}$, and a centering matrix $H = I_m - \frac{1}{m} \mathbf{1}_m \mathbf{1}_m^\top$ (where $I_m$ is the $m \times m$ identity matrix and $\mathbf{1}_m$ is an $m \times m$ matrix of ones), HSIC is calculated as: 

\begin{equation}
\text{HSIC}(K, L) = \frac{1}{(m-1)^2} \text{tr}(KHLH)
\end{equation}

The trace operator, $\text{tr}(\cdot)$, sums the diagonal elements of the resulting matrix. 
This adds the products of all corresponding entries in centered $K$ and $L$ matrices (which here are pairwise similarities between the respective network representations). 
Centering (via $H$) ensures that HSIC captures covariance structure, independent of means.  

We do not use HSIC directly as a measure of representational similarity because it is scale-dependent. 
This motivates CKA \citep{kornblith2019similarity}, which normalizes HSIC: 

\begin{equation}
    \text{CKA}(K, L) = \frac{\text{HSIC}(K, L)}{\sqrt{\text{HSIC}(K, K) \cdot \text{HSIC}(L, L)}}
\end{equation}

CKA scores range from 0 (indicating that the representations are dissimilar) to 1 (indicating that the representations are identical up to permissible transformations). 
This provides a scalar measure of how similarly two circuit heads process information across the input batch.


\subsection{Aligning Corresponding Circuit Heads}
\label{sec:distill-obj}

We have a (differentiable) measure of similarity between activations (CKA), but which pairwise similarities should we enforce between student and teacher? 
What we want is to induce a circuit in the former akin to a relevant circuit in the latter. 
Operationally this requires aligning circuit components in the teacher to analogous components in the student, i.e., finding \textit{functionally analogous} circuit heads.  
It is not obvious how to do this, particularly as student and teacher models will often differ in size (e.g., {\tt Llama8B} teacher and a {\tt Llama3B} student). 
We propose an ablation-based strategy to map student heads to teacher heads based on their respective contributions to task performance. 
This mapping then guides CKA-based alignment.

\begin{figure}
\centering
  \includegraphics[scale=.75]{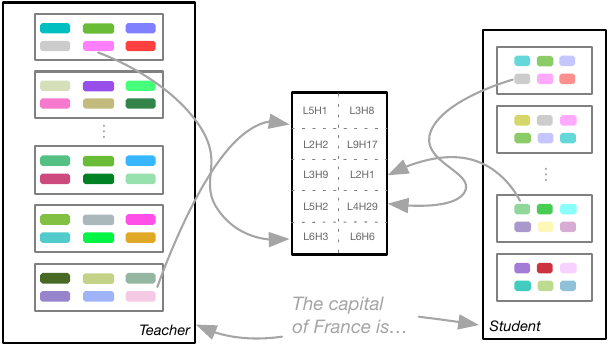}
  \caption{Circuit distillation requires mapping student heads to \textit{functionally analogous} teacher heads. Here we propose do this by comparing performance drops under ablation (other approaches may also be viable). These correspondences are then used to align model components via the composite loss during distillation.}
  \label{fig:head_alignment}
\end{figure}

The approach is as follows. 
First, we establish baseline performance metrics for both the student and teacher models on the target task dataset. 
Let {\color{purple}$P_{\text{s\_base}}$} be the performance of the student model and $P_{\text{t\_base}}$ be that of the teacher model.
We then quantify the functional importance of each relevant student head $h_s$ by recording the performance drop on its removal. 
We denote the student model's performance when head $h_s$ is ablated (i.e., its activations replaced with mean activations) by $P_{\text{s\_abl}}(h_s)$. 
The ablation impact for student head $h_s$ is then:

\begin{equation}
    \Delta P_{\text{s}}(h_s) = {\color{purple}P_{\text{s\_base}}} - P_{\text{s\_abl}}(h_s)
\end{equation}

Analogously, for each relevant teacher head ($h_t$), we record $P_{\text{t\_abl}}(h_t)$ or the teacher's  performance when head $h_t$ is ablated. 
Crucially, to align with the student's operational capabilities, the teacher head's ablation impact is calculated \textit{relative} to the student's baseline performance:

\begin{equation}
    \Delta P_{\text{t}}(h_t) = {\color{purple}P_{\text{s\_base}}} - P_{\text{t\_abl}}(h_t)
\end{equation}

This definition aims to identify teacher heads that, when ablated,  result in performance degradation comparable in magnitude to that observed when ablating a student head, relative to student performance. 
With these ablation impacts computed, we create a mapping. For each student head $h_s$, we seek teacher heads $h_t$ whose ablation impact $\Delta P_{\text{t}}(h_t)$ is most similar to $\Delta P_{\text{s}}(h_s)$. 
For this we use the absolute difference:

\begin{equation}
    d_{\text{abl}}(h_s, h_t) = |\Delta P_{\text{s}}(h_s) - \Delta P_{\text{t}}(h_t)|
\end{equation}

Smaller $d_{\text{abl}}(h_s, h_t)$ indicates a greater similarity in functional importance as measured by ablation. 
We then map $h_s$ to the $h_t$ that minimizes $d_{\text{abl}}(h_s, h_t)$. 
This yields a mapping $\mathcal{M}: \mathcal{H}_{\text{s}} \rightarrow \mathcal{P}(\mathcal{H}_{\text{t}})$, where $\mathcal{H}_{\text{s}}$ and $\mathcal{H}_{\text{t}}$ are the sets of relevant student and teacher heads, respectively. 
This mapping $\mathcal{M}$ determines the pairs of student and teacher circuit heads 
that are used in the CKA loss term of the composite loss function (described below), thereby guiding the student model to align its internal mechanisms with those functionally important counterparts identified in the teacher.

\subsection{An Objective for Circuit Distillation}
\label{sec:loss}

To align the student's circuit to the teacher's, we maximize the CKA between the corresponding heads.  
CKA values are 1 for perfect alignment (up to invariant transformations), so we define the CKA loss for a single pair of student ($K_{\text{s}}$) and teacher circuit head ($K_{\text{t}}$) Gram matrices 
as:

\begin{equation}
    \mathcal{L}_{\text{CKA}}(K_{\text{s}}, K_{\text{t}}) = 1 - \text{CKA}(K_{\text{s}}, K_{\text{t}})
    \label{eq:composite-loss}
\end{equation}

\noindent Minimizing this loss 
pushes the CKA score toward 1, i.e., towards alignment of the induced representations under the corresponding circuit heads.

We then define a composite objective for the student model as a weighted combination of the primary task loss and the sum of CKA-based circuit similarity losses over aligned circuit head pairs: 

\begin{equation}
    \mathcal{L}_{\text{total}} = {\color{blue}\mathcal{L}_{\text{task}}(y, \hat{y}_{\text{s}})} + \lambda \sum_{c \in {\color{teal}\mathcal{C}_{\text{paired}}}} \mathcal{L}_{\text{CKA}}({\color{olive}K_{\text{s}}^{(c)}}, {\color{purple}K_{\text{t}}^{(c)}})
\end{equation}

\noindent Where:
\begin{itemize}[nosep,leftmargin=*]
    \item {\color{blue}$\mathcal{L}_{\text{task}}(y, \hat{y}_{\text{s}})$} is the conventional loss for the downstream task, with $y$ denoting the teacher labels and $\hat{y}_{\text{s}}$ the student predictions.
    \item {\color{teal}$\mathcal{C}_{\text{paired}}$} denotes the set of pairs of circuit heads identified as analogous in the student and teacher models. The choice of these pairs is critical and should be informed by mechanistic understanding of which circuits in the teacher are most vital for the task or represent desirable computational properties.
    \item {\color{olive}$K_{\text{s}}^{(c)}$} and {\color{purple}$K_{\text{t}}^{(c)}$} are the Gram matrices derived from the activations of the $c$-th paired student and teacher circuit heads, respectively.
    \item $\lambda$ is a scalar hyperparameter balancing the contribution of the task performance objective against the mechanistic alignment objective. Carefully tuning $\lambda$ is necessary to ensure that the student model effectively learns the task while also internalizing the desired computational strategies from the teacher's circuits.
\end{itemize}

\section{Experimental Setup}
Next we outline our experimental design, including the models, tasks, and the methods that we use to identify relevant circuit components for our mechanistic distillation study. 
The aim of these experiments is to assess whether and to what extent circuit distillation yields stronger performing student models, compared to standard distillation (based solely on teacher outputs). 

\begin{figure*}
\centering
  \includegraphics[scale=0.3250]{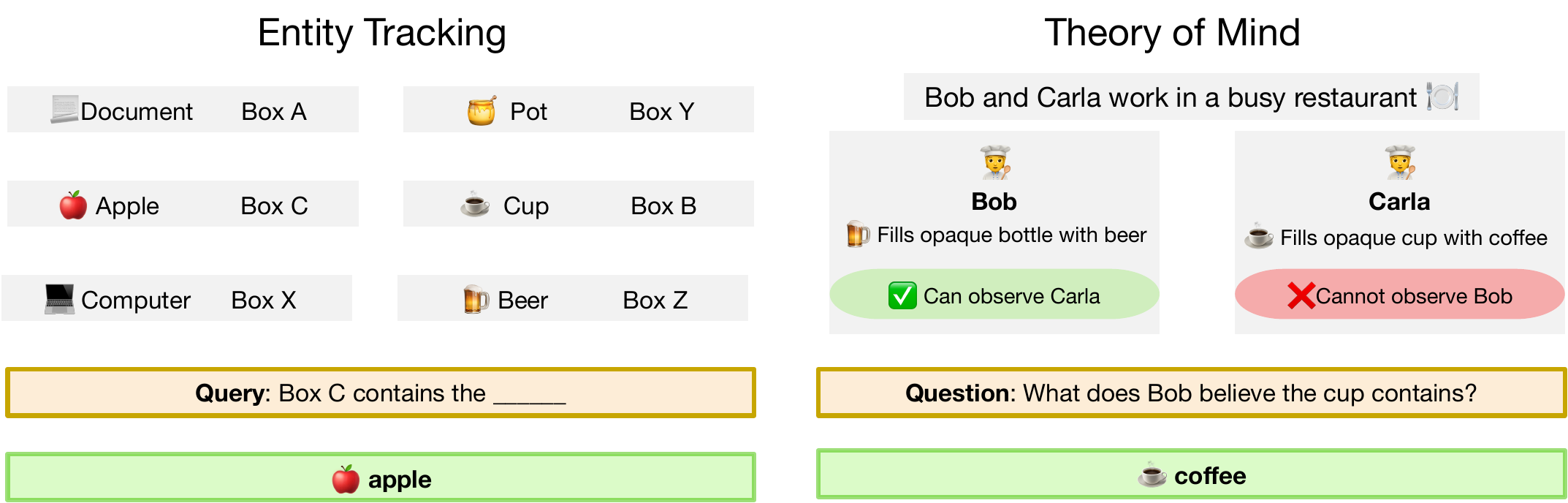}
  \caption{Illustrations of the evaluated tasks: entity tracking (left), which tests recall of item locations, and theory of mind (right), which tests reasoning about beliefs under partial observability.}
  \label{fig:task_illustration}
\end{figure*}

\subsection{Models and Tasks}
We use {\tt Llama3} models  \citep{grattafiori2024llama3herdmodels}, specifically the 1B, 3B, and 8B parameter variants. The 8B model serves as the teacher and the smaller models as students.\footnote{We use \emph{relatively} small teacher and student models throughout this work owing to compute constraints.} This setup allows us to study distillation across model capacities.

We focus on two tasks, shown schematically in Figure \ref{fig:task_illustration}. This choice is deliberate: recent mechanistic analyses have already characterized relevant circuits for both tasks \citep{prakash2024finetuningenhancesexistingmechanisms, zhu2024languagemodelsrepresentbeliefs}. Using these established circuits lets us focus on transferring a known mechanism from teacher to student, rather than on the separate and difficult problem of circuit discovery.

The first task is \textbf{Entity Tracking} (Figure \ref{fig:task_illustration}, left). Following the dataset and setup of \citet{prakash2024finetuningenhancesexistingmechanisms}, we study how models maintain and update information about entities in context. The model must identify the contents of a specific box from a preceding description of objects and boxes; for example, given ``The keys are in Box C, the phone is in Box A. Box C contains the...'', it should predict ``keys''. Motivated by prior findings that models fine-tuned on arithmetic data such as GOAT \citep{liu2023goatfinetunedllamaoutperforms} show stronger entity-tracking performance, we first fine-tune the teacher on GOAT.

The second task is causal \textbf{Theory of Mind} (ToM; Figure \ref{fig:task_illustration}, right), which probes the model’s ability to reason about beliefs under differing perspectives or access to information. We construct a dataset from prompts in \citet{prakash2025languagemodelsuselookbacks}, comprising short stories in which characters interact with objects and the model must infer beliefs from visibility and actions, e.g.: ``Sarah places her book in the drawer and leaves the room. While Sarah is gone, Tom moves the book to the shelf.'' When then asked ``Where does Sarah believe her book is?'', the model should infer Sarah's belief (the drawer) rather than the actual location (the shelf). 

For ToM, prior work found that instruction-tuned {\tt Llama3} models substantially outperform base models.\footnote{Our preliminary experiments corroborated these findings.} Because the proprietary data used for official {\tt Llama3}-Instruct models is unavailable, we prepare our teacher by instruction-tuning the larger {\tt Llama3-8B} base model on the public Alpaca dataset \citep{alpaca}. This custom instruction-tuned {\tt Llama3} model then serves as the teacher.

\begin{table*}[t]
\centering
\scriptsize
\begin{tabular}{lll|cc|cc}
\toprule
 & & & \multicolumn{2}{c|}{\textbf{Entity Tracking}} & \multicolumn{2}{c}{\textbf{Theory of Mind}} \\
\cmidrule(lr){4-5} \cmidrule(l){6-7}
 & & \textbf{Loss} & \textbf{Full Model} & \textbf{Circuit} & \textbf{Full Model} & \textbf{Circuit} \\
\midrule
\multirow{4}{*}{Baselines} & Base--1B & -- & 0.68 & 0.65 & 0.55 & 0.54 \\
 & Base--3B & -- & 0.72 & 0.70 & 0.58 & 0.56 \\
 & 8B-Instruct (Meta) & -- & -- & -- & \textbf{0.79} & \textbf{0.78} \\
 & 8B-FT$^\dagger$ & CE & \textbf{0.85} & \textbf{0.81} & 0.76 & 0.76 \\
\midrule
\multirow{4}{*}{Fully Distilled} & \multirow{2}{*}{Base--1B} & CE & 0.75 & 0.66 & 0.60 & 0.55 \\
 & & KD & 0.76 & 0.66 & 0.61 & 0.55 \\
 & \multirow{2}{*}{Base--3B} & CE & 0.78 & 0.71 & 0.63 & 0.56 \\
 & & KD & 0.78 & 0.71 & 0.63 & 0.58 \\
\midrule
\multirow{10}{*}{Circuit Distilled}
 & \multirow{5}{*}{Base--1B}
   & CE & 0.73 & 0.69 & 0.59 & 0.55 \\
 & & Align CKA & 0.68 & 0.68 & 0.52 & 0.47 \\
 & & CE + Rand CKA & 0.58 & 0.54 & 0.49 & 0.44 \\
 & & CE + Layer CKA & 0.74 & 0.71 & 0.59 & 0.59 \\
 & & \cellcolor{green!10}CE + Align CKA & \cellcolor{green!10}0.77 & \cellcolor{green!10}0.73 & \cellcolor{green!10}0.64 & \cellcolor{green!10}0.61 \\
 & \multirow{5}{*}{Base--3B}
   & CE & 0.77 & 0.75 & 0.62 & 0.57 \\
 & & Align CKA & 0.73 & 0.72 & 0.55 & 0.55 \\
 & & CE + Rand CKA & 0.63 & 0.61 & 0.49 & 0.41 \\
 & & CE + Layer CKA & 0.79 & 0.78 & 0.63 & 0.60 \\
 & & \cellcolor{green!10}CE + Align CKA & \cellcolor{green!10}\textbf{0.82} & \cellcolor{green!10}\textbf{0.79} & \cellcolor{green!10}0.65 & \cellcolor{green!10}0.65 \\
\bottomrule
\end{tabular}
\caption{Entity tracking (left) and Theory of Mind (ToM; right) results for {\tt Llama3} models. We report results for full distillation (all parameters) and circuit distillation (only matched circuit head parameters) under different losses: Cross Entropy (CE), logit distillation (KD; \citealt{hinton2015distillingknowledgeneuralnetwork}), Centered Kernel Alignment (CKA), CKA with Random Head Assignment, CE + CKA between whole-layer attention outputs at the layers containing circuit heads (Layer CKA; no head mapping), and CE + CKA with ablation-based head alignment loss. $^\dagger$8B-FT is GOAT-8B for entity tracking and 8B-Alpaca for ToM. Chance accuracy on entity tracking is 0.14. Instruction-tuned models (e.g., Llama3-Instruct) outperform their base counterparts on ToM. On both tasks, the improvements of circuit distillation (1B, 3B) using CE + Align CKA are statistically significantly better than standard full distillation under CE and circuit distillation using CE only ($p$$<$$0.05$) under McNemar's test (details in Appendix).}
\label{tab:res_et}
\end{table*}

\subsection{Identification of Relevant Circuit Heads}
\label{sec:circuit_discovery}
A key component of our proposed mechanistic distillation approach is targeting specific (teacher) circuit heads relevant to a given task. 
Methods for identifying such circuits are currently task-dependent, and automated circuit discovery is an active area of research \citep{conmy2023towards,wang2022interpretabilitywildcircuitindirect}. 
Here we rely on prior efforts that have characterized circuits for the two tasks we consider. 

For entity tracking, we follow path patching as described in \cite{prakash2024finetuningenhancesexistingmechanisms}.
This identifies a sparse set of attention heads that form the core circuit responsible for tracking entities and their properties. 
For the ToM task, where distinct belief-tracking and reasoning circuits are hypothesized, we use activation patching (with mean ablation) to assess causal impact of individual attention heads to identify the top-$n$ heads most critical for successful ToM reasoning. 
Details of these experiments and patching methods are provided in Appendix \ref{appx:exp_details}.

The sets of circuit heads identified through these procedures for both tasks subsequently serve as alignment targets in our CKA-based mechanistic distillation objective described in Section \ref{sec:methods}.

\section{Results}
This section presents the empirical validation of our mechanistic distillation framework on the two tasks introduced above: Entity Tracking and ToM. 
We first identify relevant teacher circuits based on prior work (Section \ref{sec:circuit_discovery}), then search for corresponding student circuits as described in Section \ref{sec:distill-obj}. 
We compare standard behavioral distillation (with both full-model and targeted-circuit updates; CE only) against our circuit distillation method (CE + Align CKA) and a key control condition (CE + Rand CKA) that isolates the effect of intentional circuit alignment. 
We also compare against logit distillation and a layer-level alignment baseline (Section \ref{sec:layer_vs_circuit}), and test directly whether teacher circuits are  transferred (Section \ref{sec:mech_transfer}).\footnote{We note that circuit distillation also affords efficiency gains: See Appendix Figure \ref{fig:appx_lossplot}.}


\subsection{Entity Tracking}
Table \ref{tab:res_et} (left) summarizes results on the entity tracking task and shows the efficacy of aligning internal mechanisms. The ``Full Model'' column reports accuracy using the entire model, while ``Circuit'' reports performance when only the pre-identified entity tracking circuit is active and all other attention heads are mean-ablated.

The teacher achieves 0.85 accuracy, and its identified entity tracking circuit, operating in isolation, nearly matches this at 0.81. This high fidelity indicates that the discovered circuit is the primary locus of the teacher’s entity tracking capability, making it a suitable target for distillation.

The base {\tt Llama3-3B} student starts at 0.72 accuracy, leaving a substantial gap to close. Traditional distillation improves performance to 0.78 with full-model updates and 0.77 with targeted circuit updates, which involve only about 11\% of attention heads. The smaller gap between full-model and circuit-only performance in the latter case suggests that the proposed approach can identify relevant student circuits, and that standard CE loss can transfer circuits to some extent when training is restricted to circuit-relevant parameters.

Circuit distillation (\texttt{CE + Align CKA}) applies the composite loss from Equation \ref{eq:composite-loss} to functionally aligned heads and reaches 0.82 accuracy, with 0.79 for the isolated circuit. 
This is a substantial improvement over CE alone and nearly closes the teacher-student gap in circuit-only performance.

To assess the importance of explicit \textit{alignment of circuit components}, we also apply CKA after randomly assigning each identified teacher head to a student head \texttt{(CE + Rand CKA)}. This reduces accuracy to 0.63, below the CE baseline, showing that CKA is not helpful when applied indiscriminately. 

Taken together, these results suggest that enforcing representational alignment on a small, functionally critical subset of components enables more direct and effective transfer of the underlying computational mechanisms than matching the teacher’s final predictions alone.

\subsection{Theory of Mind}
To assess whether our framework generalizes beyond retrieval-based tasks to more complex reasoning problems, we apply the same methodology to the ToM task. The results in Table \ref{tab:res_et} (right) corroborate and extend our initial findings.

As a performance ceiling, we evaluate \texttt{Llama3-8B-Instruct}, which achieves 0.79 accuracy. 
Recall from above that we fine-tuned our own version using the Alpaca data; this \texttt{8B-Alpaca} model reaches 0.76 accuracy. 
The base {\tt Llama3-3B} student shows limited ToM ability: 0.58 accuracy;  0.56 using only its identified circuit. 

Our main experiments (Table \ref{tab:res_et}, bottom row), where we train only the identified ToM circuit heads, quantify component contributions. 
Behavioral distillation using only Cross-Entropy loss \texttt{(CE only)} raises full-model accuracy to 0.62 and circuit accuracy to 0.57. 
Applying a CKA loss with randomly mapped heads \texttt{(CE + Rand CKA)} \emph{reduces} performance to 0.49 (full model) and 0.41 (circuit), below the student baseline; alignment between functionally irrelevant components is harmful. 
Circuit distillation \texttt{(CE + Align CKA)} achieves 0.65 accuracy for both the full model and the isolated circuit, indicating that the gains come from the targeted circuit and supporting the effectiveness of our functional alignment strategy.

To assess statistical significance, we use McNemar’s tests on an evaluation set of ${N = 1000}$ samples. Specifically, we compare the CE baseline against our proposed CE + Align CKA variant, and likewise compare circuit-distilled models trained with CE only versus CE + Align CKA. All comparisons are significant ($p < .05$).\footnote{Details in Appendix Section \ref{appx_seca1}.}

\subsection{Is Circuit-Level Alignment Necessary?}
\label{sec:layer_vs_circuit}
While the CE + Rand CKA control shows that aligning 
\emph{random} heads is harmful, other kinds of representation matching may help. 
We add two baselines to Table \ref{tab:res_et}. \textbf{CE + Layer CKA} applies the same CKA objective to the full attention output of every student layer that contains a circuit head, paired with the teacher layer at the same relative depth. It uses no head mapping and updates the same student parameters as circuit distillation, isolating the choice of alignment \emph{target}. \textbf{KD} is standard logit distillation \citep{hinton2015distillingknowledgeneuralnetwork}: all student parameters are trained to match the teacher's temperature-scaled output distribution ($T = 2$).

On entity tracking, layer-level alignment reaches \textbf{0.74} (1B) and \textbf{0.79} (3B), compared to 0.77 and 0.82 for CE + Align CKA, and KD reaches 0.76 and 0.78. On ToM, the corresponding numbers are 0.59/0.63 (Layer CKA), 0.64/0.65 (CE + Align CKA), and 0.61/0.63 (KD). Aligning the residual stream at the same layers instead of the attention output performs similarly or slightly worse (0.71/0.78 on entity tracking). Together with the Rand CKA control, these results order the alignment targets as matched circuit heads $>$ whole layers $>$ mismatched heads, suggesting that the gains depend on head-level functional correspondence rather than on representation matching per se.

\begin{figure*}[t]
\centering
  \includegraphics[scale=.8]{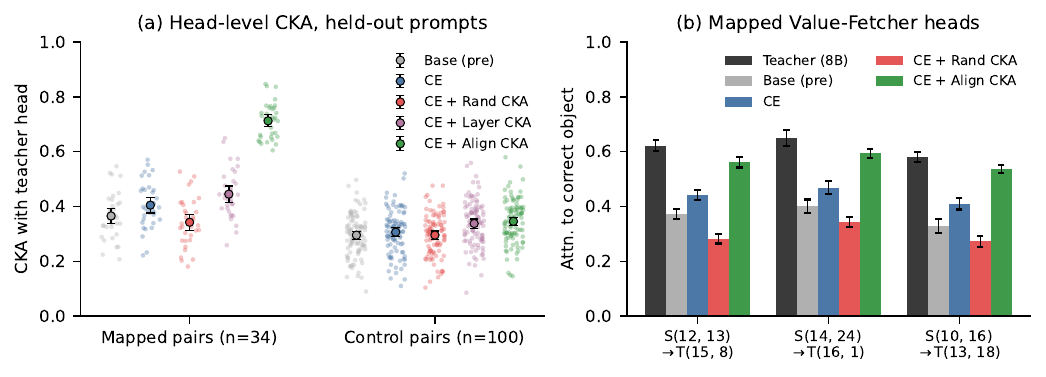}
  \caption{Mechanism transfer on entity tracking ({\tt Llama3-1B} student, 8B teacher), measured on held-out prompts. (a) CKA between each student circuit head and its mapped teacher head (left) and between the same student heads and non-mapped teacher circuit heads (control, right), before (Base) and after distillation; large markers show means with 95\% confidence intervals. (b) Attention from the final token to the correct object token for the three mapped pairs with the strongest teacher Value Fetcher scores; error bars are standard errors.}
  \label{fig:mech_verify}
\end{figure*}

\begin{table*}[t]
\centering
\scriptsize
\begin{tabular}{lccc}
\toprule
\textbf{Student} & \textbf{VF overlap@20} & \textbf{Spearman $\rho$ (S vs.\ T)} & \textbf{Mean mapped PP score} \\
\midrule
Base (pre) & 1.00 & 0.18 & $-$0.021 \\
CE & 0.80 & 0.27 & $-$0.026 \\
\cellcolor{green!10}CE + Align CKA & \cellcolor{green!10}0.90 & \cellcolor{green!10} 0.55 & \cellcolor{green!10} $-$0.041 \\
\bottomrule
\end{tabular}
\caption{Path patching (Value Fetcher stage) re-run on distilled {\tt Llama3-1B} students. \emph{VF overlap@20}: fraction of the 20 most causally important heads at the final position that belong to the student circuit. \emph{Spearman $\rho$}: rank correlation between the patching scores of mapped student heads and those of their teacher counterparts. \emph{Mean mapped PP score}: mean patching score of mapped student heads (more negative = more causally important).}
\label{tab:pp_verify}
\end{table*}

\subsection{Is the Mechanism Transferred?}
\label{sec:mech_transfer}
Higher task accuracy and higher circuit-only accuracy are consistent with mechanism transfer, but do not necessarily establish mechanism transfer. 
We test this directly on the entity tracking students ({\tt Llama3-1B}), using held-out prompts not seen during distillation (Figure \ref{fig:mech_verify}, Table \ref{tab:pp_verify}).

\paragraph{Representational alignment.} For each mapped pair, we compute CKA between the student head and its teacher counterpart before and after training (Figure \ref{fig:mech_verify}a). Training with CE alone changes mapped-pair CKA only marginally (0.37 $\rightarrow$ 0.41), and layer-level alignment raises it only modestly (0.45), whereas CE + Align CKA raises it to 0.71. The increase is specific to the aligned pairs: CKA between the same student heads and \emph{non-mapped} teacher circuit heads stays at 0.30--0.35, and CE + Rand CKA leaves the ablation-mapped pairs unchanged (0.34).

\paragraph{Attention behavior.} In the teacher, Value Fetcher heads attend from the final token to the object stored in the queried box \citep{prakash2024finetuningenhancesexistingmechanisms}. For the three mapped pairs with the strongest teacher Value Fetcher scores, we measure the attention that the corresponding student head places on that object token (Figure \ref{fig:mech_verify}b). After circuit distillation, student heads place 0.56 of their attention on the correct object, as compared to 0.37 in the base student, and 0.44 after CE-only training.

\paragraph{Causal structure.} Finally, we re-run the first stage of path patching (Value Fetcher identification; Appendix \ref{appx:circuit_id}) on the distilled students (Table \ref{tab:pp_verify}). After CE + Align CKA, 18 of the 20 most causally important heads at the final position are in the trained circuit (16 after CE alone), and the rank correlation between the patching scores of mapped student heads and their teacher counterparts rises from 0.18 before training to 0.55 (0.27 after CE alone), i.e., the aligned student heads come to play the same causal roles as their teacher counterparts, which supports mechanism transfer rather than a generic regularization effect of the CKA term.

\section{Related Work}
This work connects four active areas: knowledge distillation, mechanistic interpretability, causal abstraction, and circuit analysis in language models.

\vspace{.35em}
\noindent {\bf Knowledge Distillation}
Knowledge distillation was introduced as model compression, training a smaller student to match a larger teacher’s soft outputs \citep{hinton2015distillingknowledgeneuralnetwork}. More recent work uses teacher models to generate instruction-following data for smaller models \citep{alpaca, chung2022scalinginstructionfinetunedlanguagemodels}, and some methods distill reasoning traces such as Chain-of-Thought or intermediate representations \citep{wei2023chainofthoughtpromptingelicitsreasoning, li2024symbolicchainofthoughtdistillationsmall, park2024selfknowledgedistillationlearningambiguity}. Our work differs from these behavioral or full-layer approaches by aiming to transfer a known computational mechanism through alignment of specific circuits.

\vspace{0.35em}
\noindent {\bf{Mechanistic Interpretability}}
Mechanistic interpretability seeks to reverse-engineer neural networks into human-understandable algorithms \citep{olah2018the}. Central to this effort is the notion of a circuit: a localized subgraph implementing a scrutable computation \citep{elhage2021mathematical, wang2022interpretabilitywildcircuitindirect}. Such circuits are typically identified with causal methods like path or activation patching \citep{goldowskydill2023localizingmodelbehaviorpath, conmy2023automatedcircuitdiscoverymechanistic}, and have been used to explain behaviors ranging from factual recall to indirect object identification \citep{meng2023locatingeditingfactualassociations, wang2022interpretabilitywildcircuitindirect}. These results suggest that specialized subgraphs are meaningful targets for transfer.

\vspace{0.35em}
\noindent {\bf{Causal Abstraction and Alignment}}
Causal abstraction provides a formal framework for testing algorithmic equivalence between models \citep{10.5555/3540261.3540994, Geiger2023CausalAA, sun2025hyperdasautomatingmechanisticinterpretability}, and prior work has explored alignment for circuit discovery \citep{gupta2025interpbenchsemisynthetictransformersevaluating}. 
Unlike these verification-oriented approaches, our goal is distillation: we use a CKA-based objective to transfer targeted capabilities rather than prove strict equivalence.

\vspace{0.35em}
\noindent {\bf Circuits for Cognitive Phenomena}
Circuit methods have been especially useful for studying cognitive phenomena such as entity tracking and theory of mind. Prior work has examined entity binding and tracking \citep{feng2024languagemodelsbindentities, li-etal-2021-implicit}, and \citet{prakash2024finetuningenhancesexistingmechanisms} identified a sparse circuit for entity tracking in Llama-family models. Theory-of-mind abilities have been widely studied behaviorally \citep{shapira-etal-2024-clever, Kosinski_2024, le-etal-2019-revisiting, xu-etal-2024-opentom, 95913661aa554a069d211d907aa3a2bb, chan-etal-2024-negotiationtom, jin2024mmtomqa}, while recent mechanistic work has begun identifying underlying computations such as the “lookback mechanism” for belief tracking \citep{prakash2025languagemodelsuselookbacks}. 
Our work is about transferring known mechanisms rather than discovering new ones; 
to our knowledge, this is the first work to frame distillation as targeted mechanistic transfer.

\section{Conclusions}
We have proposed \emph{circuit distillation}, a mechanistic alternative to traditional model distillation that aims to instill relevant internal algorithms (\emph{circuits}) in a student model, rather than merely mimic output behavior. Our method first identifies functionally analogous circuit components in teacher and student models using ablation-impact similarity, then enforces mechanistic alignment with a Centered Kernel Alignment (CKA)-based loss. 

On two cognitive tasks---entity tracking and theory of mind---this approach outperforms standard behavioral distillation, while updating only 11--15\% of the student model's attention heads. These results suggest that direct circuit alignment can be a more effective strategy for algorithmic transfer than output mimicry alone, offering both a useful form of model compression and a route to more interpretable student models.

\section*{Limitations}
This work does have important limitations. It depends on the prior identification of a well-defined teacher circuit, which remains difficult and is still an active area of mechanistic interpretability research. Our ablation-based mapping of teacher to student components is heuristic, and more sophisticated approaches to functional correspondence across architectures may yield improvements. Our empirical analysis is also limited to a small number of tasks with well-defined circuits and to relatively small models due to compute constraints. For practitioners who want a generally capable small model, standard distillation remains the right tool. Circuit distillation is complementary: it is suited to instilling a targeted, well-characterized capability, potentially on top of a student that has already been distilled with standard methods. Even so, we believe this work points to promising directions for transferring targeted functionality beyond output mimicry.


\bibliography{custom}
\clearpage

\appendix
\begin{table*}
\centering
\scriptsize
\begin{tabular}{lllcccc}
\toprule
\textbf{} & \textbf{} & \textbf{Loss} & 
\textbf{Full Model} & \textbf{Circuit} & \textbf{Random Circuit} & \textbf{Faithfulness} \\
\midrule
\multirow{6}{*}{Baselines} & Base-1B & -- & 0.68 & 0.65 & 0.03 & 0.96 \\
& Base-3B & -- & 0.72 & 0.70 & 0.02 & 0.97 \\
& Base-8B & -- & 0.73 & 0.72 & 0.01 & 0.97 \\
& GOAT-1B & CE & 0.77 & 0.69 & 0.03 & 0.89 \\
& GOAT-3B & CE & 0.83 & 0.75 & 0.04 & 0.90 \\
& GOAT-8B& CE & 0.85 & 0.81 & 0.02 & 0.89 \\
\midrule
\multirow{4}{*}{Fully Distilled} & \multirow{2}{*}{Base-1B}  & CE & 0.75 & 0.66 & 0.01 & 0.88 \\
 & & KD & 0.76 & 0.66 & 0.02 & 0.88 \\
& \multirow{2}{*}{Base-3B} & CE & 0.78 & 0.71 & 0.03 & 0.91 \\
 & & KD & 0.78 & 0.71 & 0.02 & 0.91 \\
\midrule
\multirow{24}{*}{Circuit Distilled} 
 & \multirow{6}{*}{Base--1B (GOAT-style)} 
   & CE & 0.71 & 0.69 & 0.01 & 0.97 \\
 & & Align CKA & 0.69 & 0.57 & 0.01 & 0.82 \\
 & & CKA + Rand CKA & 0.62 & 0.55 & 0.02 & 0.88 \\
 & & CE + Layer CKA (attn) & 0.73 & 0.71 & 0.02 & 0.97 \\
 & & CE + Layer CKA (resid) & 0.72 & 0.70 & 0.01 & 0.97 \\
 & & \cellcolor{green!10}CE + Align CKA & \cellcolor{green!10}0.75 & \cellcolor{green!10}0.74 & \cellcolor{green!10}0.03 & \cellcolor{green!10}0.98 \\
 & \multirow{6}{*}{Base--1B} 
   & CE & 0.73 & 0.69 & 0.02 & 0.94 \\
 & & Align CKA & 0.68 & 0.68 & 0.01 & 1.00 \\
 & & CE + Rand CKA & 0.58 & 0.54 & 0.01 & 0.93 \\
 & & CE + Layer CKA (attn) & 0.75 & 0.71 & 0.02 & 0.95 \\
 & & CE + Layer CKA (resid) & 0.74 & 0.71 & 0.01 & 0.95 \\
 & & \cellcolor{green!10}CE + Align CKA & \cellcolor{green!10}0.77 & \cellcolor{green!10}0.73 & \cellcolor{green!10}0.01 & \cellcolor{green!10}0.94 \\
 & \multirow{6}{*}{Base--3B (GOAT-style)} 
   & CE & 0.76 & 0.71 & 0.02 & 0.95 \\
 & & Align CKA & 0.72 & 0.71 & 0.02 & 0.95 \\
 & & CKA + Rand CKA & 0.69 & 0.62 & 0.01 & 0.90 \\
 & & CE + Layer CKA (attn) & 0.76 & 0.73 & 0.02 & 0.96 \\
 & & CE + Layer CKA (resid) & 0.76 & 0.73 & 0.01 & 0.96 \\
 & & \cellcolor{green!10}CE + Align CKA & \cellcolor{green!10}0.77 & \cellcolor{green!10}0.77 & \cellcolor{green!10}0.01 & \cellcolor{green!10}1.00 \\
 & \multirow{6}{*}{Base--3B} 
   & CE & 0.77 & 0.75 & 0.02 & 0.98 \\
 & & Align CKA & 0.73 & 0.72 & 0.00 & 0.99 \\
 & & CKA + Rand CKA & 0.63 & 0.61 & 0.01 & 0.99 \\
 & & CE + Layer CKA (attn) & 0.79 & 0.77 & 0.02 & 0.97 \\
 & & CE + Layer CKA (resid) & 0.79 & 0.78 & 0.01 & 0.96 \\
 & & \cellcolor{green!10}CE + Align CKA & \cellcolor{green!10}\textbf{0.82} & \cellcolor{green!10}\textbf{0.79} & \cellcolor{green!10}0.00 & \cellcolor{green!10}0.98 \\
\bottomrule
\end{tabular}
\caption{Extended results on the Entity Tracking task.}
\label{tab:et_extended_res}
\end{table*}

\begin{table*}
\centering
\scriptsize
\begin{tabular}{lllcccc}
\toprule
\textbf{} & \textbf{} & \textbf{Loss} & 
\textbf{Full Model} & \textbf{Circuit} & \textbf{Random Circuit} & \textbf{Faithfulness} \\
\midrule
\multirow{9}{*}{Baselines}  & Base-1B & -- & 0.55 & 0.54 & 0.02 & 0.98 \\
& Base-3B & -- & 0.58 & 0.56 & 0.01 & 0.97 \\
& Base-8B & -- & 0.71 & 0.68 & 0.02 & 0.95 \\
& Base-1B-Instruct & -- & 0.65 & 0.64 & 0.02 & 0.98 \\
& Base-3B-Instruct & -- & 0.69 & 0.67 & 0.01 & 0.97 \\
& Base-8B-Instruct & -- & 0.79 & 0.78 & 0.02 & 0.99 \\
& Base-1B-Alpaca & CE & 0.58 & 0.53 & 0.03 & 0.91 \\
& Base-3B-Alpaca & CE & 0.66 & 0.64 & 0.05 & 0.97 \\
& Base-8B-Alpaca & CE & 0.76 & 0.0.71 & 0.02 & 0.93 \\
\midrule
\multirow{4}{*}{FullyDistilled}  & \multirow{2}{*}{Base-1B} & CE & 0.60 & 0.55 & 0.02 & 0.92 \\
 & & KD & 0.61 & 0.55 & 0.02 & 0.92 \\
& \multirow{2}{*}{Base-3B} & CE & 0.63 & 0.56 & 0.01 & 0.89 \\
 & & KD & 0.64 & 0.57 & 0.01 & 0.89 \\
\midrule
\multirow{12}{*}{CircuitDistilled} 
 & \multirow{6}{*}{Base 1B} 
   & CE & 0.59 & 0.55 & 0.01 & 0.93 \\
 & & Align CKA & 0.52 & 0.47 & 0.01 & 0.90 \\
 & & CKA + Rand CKA & 0.49 & 0.44 & 0.04 & 0.89 \\
 & & CE + Layer CKA (attn) & 0.61 & 0.57 & 0.02 & 0.93 \\
 & & CE + Layer CKA (resid) & 0.59 & 0.59 & 0.01 & 0.95 \\
 & & \cellcolor{green!10}CE + Align CKA & \cellcolor{green!10}0.64 & \cellcolor{green!10}0.61 & \cellcolor{green!10}0.03 & \cellcolor{green!10}0.95 \\
 & \multirow{6}{*}{Base 3B} 
   & CE & 0.62 & 0.57 & 0.00 & 0.92 \\
 & & Align CKA & 0.55 & 0.55 & 0.00 & 1.00 \\
 & & CE + Rand CKA & 0.49 & 0.41 & 0.01 & 0.84 \\
 & & CE + Layer CKA (attn) & 0.63 & 0.60 & 0.02 & 0.95 \\
 & & CE + Layer CKA (resid) & 0.63 & 0.60 & 0.01 & 0.94 \\
 & & \cellcolor{green!10}CE + Align CKA & \cellcolor{green!10}0.65 & \cellcolor{green!10}0.64 & \cellcolor{green!10}0.01 & \cellcolor{green!10}0.98 \\
\bottomrule
\end{tabular}
\caption{Extended results on the ToM task.}
\label{tab:tom_extended_res}
\end{table*}

\section{Additional Results}
Tables \ref{tab:et_extended_res} and \ref{tab:tom_extended_res} report additional comprehensive set of results, offering a granular analysis across different model scales and training paradigms. 
These tables include several key baselines and metrics designed to comprehensively evaluate our mechanistic distillation approach and validate the underlying circuits.

First, we establish a clear performance landscape. The {\tt Base-1B/3B/8B} models represent the innate capabilities of the open-source {\tt Llama3} family on these tasks. The fully fine-tuned versions on GOAT and Alpaca serve as a practical upper bound on entity tracking and theory of mind, respectively. They demonstrate the performance achievable with direct, supervised training on the target data, against which we can measure the efficacy of our distillation approaches. For example, the {\tt GOAT-8B} model achieves an accuracy of 0.85 on entity tracking. 

Before evaluating distillation, we validate the integrity of the circuits themselves using two crucial metrics. The \textbf{Faithfulness} score, calculated as the ratio of {\tt Circuit Accuracy} to {\tt Full Model Accuracy}, quantifies how much of a model's capability is captured by its identified circuit. Across all base and fine-tuned models, faithfulness remains high (0.89-0.97), confirming that the identified circuits are indeed the primary mechanisms responsible for underlying tasks. As a sanity check, the \textbf{Random Circuit} column reports the performance of a randomly selected set of model components of the same size. The near-zero accuracy (0.01-0.04) in all cases provides strong evidence that identified circuits are non-trivial, functionally cohesive units and not arbitrary collections of components.

The central comparison in our study is between traditional behavioral distillation and our mechanistic approach. The {\tt Distilled-1B} and {\tt Distilled-3B} models represent the former, where the entire student model is fine-tuned on the teacher's outputs using a standard Cross-Entropy (CE) loss. These models show solid improvement over their base counterparts (e.g. {\tt Distilled-3B} reaching 0.78 accuracy on the entity tracking task), but the {\tt Circuit Distilled} models consistently outperform this baseline. 

The extended tables also include the two baselines introduced in Section \ref{sec:layer_vs_circuit}. {\tt KD} trains all student parameters with logit distillation, and {\tt CE + Layer CKA} aligns whole-layer representations at the layers containing circuit heads, either the attention output used in the main text ({\tt attn}) or the residual stream ({\tt resid}). Across all student blocks, both Layer CKA variants improve over {\tt CE} but fall short of {\tt CE + Align CKA}, with the residual-stream variant slightly weaker than the attention-output variant. Faithfulness stays high for both variants (0.93--0.97), so the identified circuit remains the locus of the task under layer-level alignment as well.

\label{appx_seca1}
\subsection{Statistical significance testing}
To verify that the observed performance improvements of our framework are statistically significant, we use McNemar's test. This non-parametric test is well-suited for comparing the predictions of two models on the same test set by analyzing their disagreements in a contingency table.

For each {\tt Circuit Distilled} model (1B and 3B), we ran two separate significance tests:

\begin{enumerate}
    \item We compared the predictions of our proposed method ({\tt CE + Align CKA}) against the circuit-only behavioral baseline ({\tt CE} in the {\tt Circuit Distilled} block). This test assesses whether the addition of the CKA alignment term provides a significant benefit over standard training for the circuit.
    \item We compared the predictions of {\tt CE + Align CKA} against the full-model behavioral distillation baseline ({\tt Distilled} with {\tt CE} loss). This more stringent comparison evaluates whether training only the circuit mechanistically can significantly outperform training the entire model behaviorally.
\end{enumerate}

In all comparisons across both the 1B and 3B student models, the resulting $p$-values were well below the standard significance threshold of $\alpha = 0.05$, ranging from 0.003 to 0.007, providing strong statistical evidence that the performance gains achieved by our mechanistic distillation framework are not attributable to random chance and represent a genuine improvement over the baseline methods.

We additionally compare {\tt CE + Align CKA} against {\tt CE + Layer CKA} ({\tt attn}), evaluating on the full held-out entity tracking set ($N = 3{,}080$) and the ToM evaluation set ($N = 1{,}000$). The difference is significant on entity tracking ($p = 0.031$ for 1B and $p = 0.004$ for 3B) and for the 1B ToM student ($p = 0.042$), but not for the 3B ToM student ($p = 0.19$), where the full-model gap is only 0.02.

\begin{figure*}
\centering
  \includegraphics[scale=0.495]{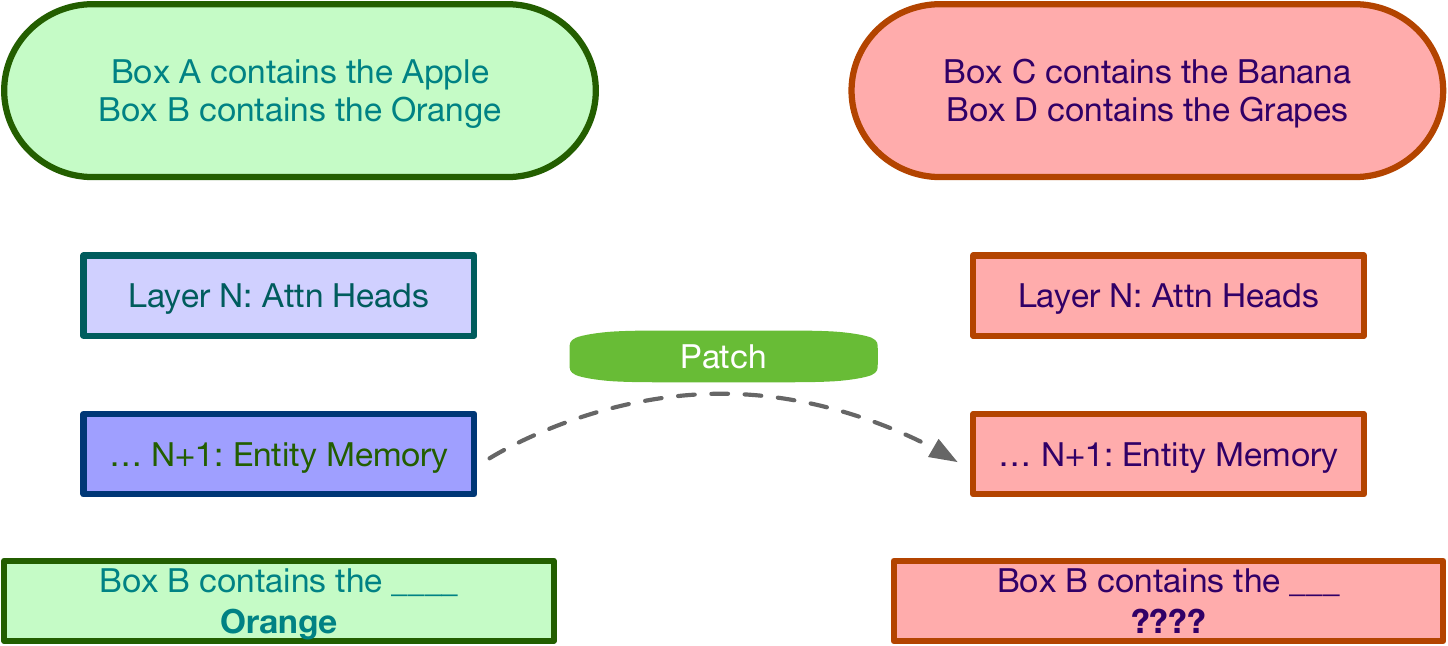}
  \caption{Path patching isolates which model components store and retrieve entity-attribute bindings.}
  \label{fig:appx_pp}
\end{figure*}

\section{Circuit Identification and Model Selection}
\label{appx:circuit_id}
Here we provide details on the methodology used to identify the entity tracking circuits that serve as the primary targets for our mechanistic distillation experiments. For this task, our approach is a direct adaptation of the Path Patching technique described by \cite{prakash2024finetuningenhancesexistingmechanisms}, which has been demonstrated to effectively discover sparse, causal circuits for entity tracking in Llama-family models. Path Patching is an iterative, causal-tracing method designed to identify the subgraph of model components---in this case, attention heads---responsible for a specific behavior. The core of the technique involves comparing two forward passes: A {\tt clean run} on an original, unmodified task input, and a {\tt corrupted run} on a counterfactual input where the information required to produce the correct answer has been removed or altered. For the entity tracking task, a corrupted input is generated by randomizing the query box, the object names, and their corresponding box labels, thus ensuring that the model cannot solve the task through simple heuristics.

The process aims to find paths in the computational graph that, when ``patched'' with activations from the corrupted run, cause the most significant degradation in the clean run's performance. 
A ``patch'' involves replacing the activation of a specific node (e.g., an attention head at a given token position) in the clean run with the corresponding activation from the corrupted run. The causal importance of each path is quantified by a patching score, which measures the relative change in the log-probability of the correct output token when that path is patched. Paths yielding the largest performance drop (i.e., the most negative scores) are considered the most causally significant and are iteratively added to the identified circuit. This iterative search begins at the final token position to identify heads that directly influence the output logit (referred to as "Value Fetcher" heads in \cite{prakash2024finetuningenhancesexistingmechanisms}). The search then proceeds backward through the model, identifying upstream heads that provide crucial information to the already-identified circuit components via query-key or value-vector compositions. This process reveals a sparse, multi-component circuit responsible for locating the query, transmitting positional information, and ultimately retrieving the correct entity value.

\paragraph{Model selection} Our decision to use GOAT-finetuned models as teachers for the entity tracking task is directly informed by the findings that emerge from this circuit analysis. \citet{prakash2024finetuningenhancesexistingmechanisms} demonstrated that while a base {\tt Llama} model possesses a nascent entity tracking circuit, fine-tuning on structured data like the arithmetic expressions in the GOAT dataset significantly enhances the functionality of this \textit{same} underlying circuit. Specifically, the {\tt Value Fetcher} and {\tt Position Transmitter} components of the circuit become more precise and effective at resolving positional information and retrieving the correct object value. This finding provides a strong mechanistic basis for our experimental design: by selecting GOAT-finetuned models as teachers, we are targeting a known, more potent version of the very mechanism we aim to instill in the student model through distillation.

\section{Experimental Details}
\label{appx:exp_details}
All experiments were conducted on a single NVIDIA A100 GPU. To maintain experimental control and isolate the effects of the different loss functions, we used the default hyperparameters of the base {\tt Llama3} models for all training runs. The only modification was a consistent learning rate of 2e-5, which was applied across all teacher fine-tuning, standard distillation, and mechanistic distillation experiments.
When training circuit heads only, we mask gradients so that only the query and output-projection slices of the circuit heads, and the key/value slices of their grouped-query attention groups, are updated; all other parameters stay frozen. We use $\lambda = 0.5$ for all CKA-based objectives (Align CKA, Rand CKA, Layer CKA) and temperature $T = 2$ for KD.

The datasets for our two tasks were generated as follows:

\paragraph{Entity Tracking:} The dataset was created by \citet{kim-schuster-2023-entity}, where they designed it to evaluate a model's ability to track the state changes of discourse entities. The dataset contains English sentences describing a setting where various objects are located in different boxes, and the task is to predict the contents of a queried box. 

Following  \citet{prakash2024finetuningenhancesexistingmechanisms}, we modify the structure of the context segment. Instead of the original format ``Box F contains the apple,'' we reorder the phrases to ``The apple is in box F.'' This structural change is crucial as it prevents the model from simply relying on shallow pattern matching or locating the longest identical context segment between the context and the query. It forces the model to genuinely infer the relationship between an object and its container. Each task instance in our setup involves several boxes, each labeled with a random letter and containing a unique, single-token object. 

\paragraph{Theory of Mind (ToM)} For the ToM task, we use the CausalToM dataset \citep{prakash2025languagemodelsuselookbacks}. 
This dataset was specifically constructed for the causal analysis of ToM reasoning, addressing the limitations of existing datasets which often lack the structure needed for counterfactual interventions. Each story involves two characters who each interact with a distinct object, causing it to take on a unique state (e.g., ``Carla grabs an opaque cup and fills it with coffee. Then Bob grabs another opaque bottle and fills it with beer.''). The model is then asked to reason about one character's belief regarding an object's state, often under conditions of limited or no visibility of the other character's actions. 

\begin{figure*}
\centering
  \includegraphics[scale=0.384]{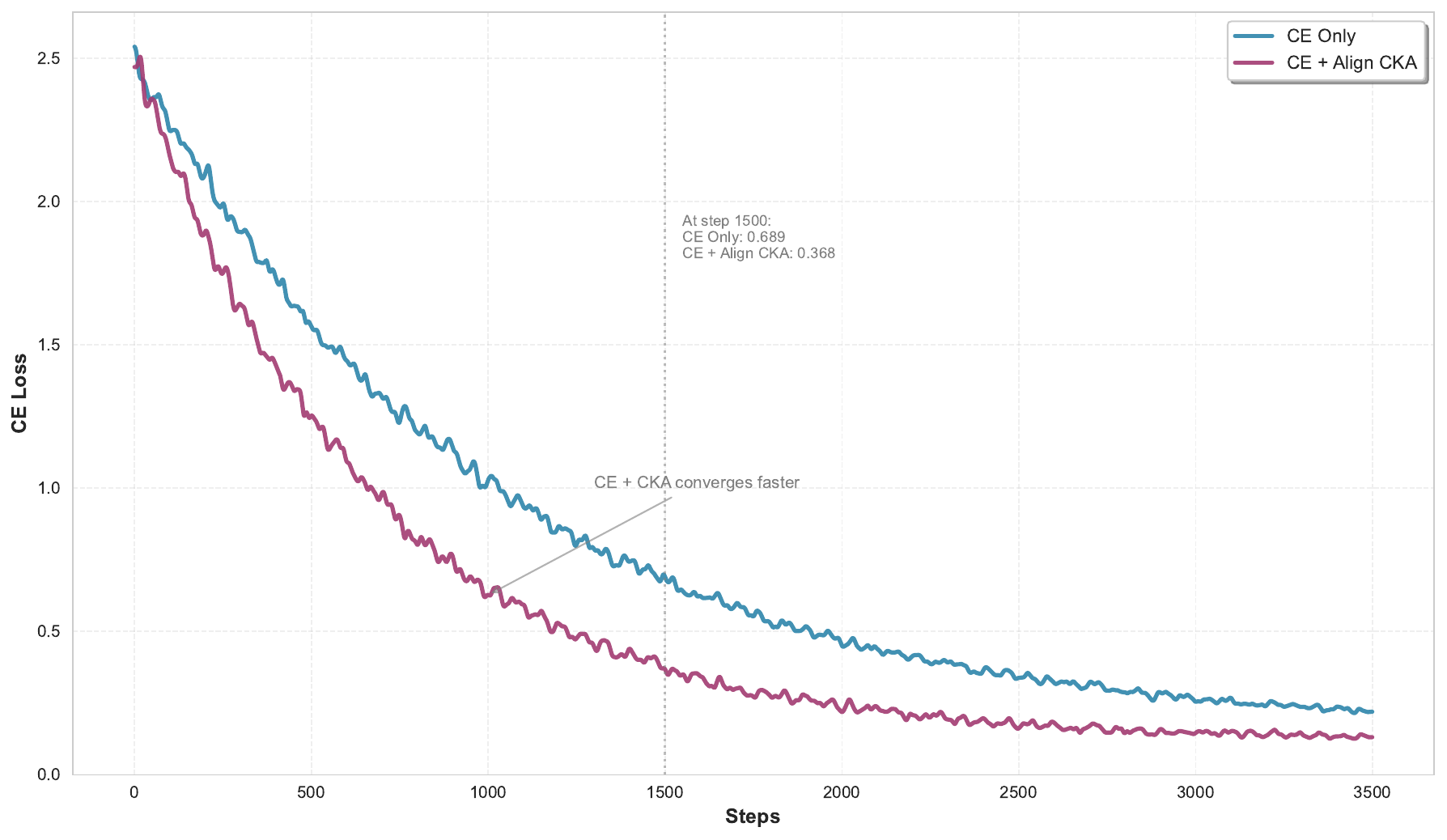}
  \caption{We compare the Cross-Entropy (CE) loss component for the {\tt CircuitDistilled-3B} model on the entity tracking task when trained with a standard behavioral objective ({\color{cyan}CE Only}) versus our mechanistic distillation method ({\color{violet}CE + Align CKA})}.
  \label{fig:appx_lossplot}
\end{figure*}

\subsection{Training dynamics and convergence}
An analysis of the training dynamics reveals that our mechanistic distillation approach not only achieves higher final accuracy but also learns more efficiently. 

Figure \ref{fig:appx_lossplot} plots the cross entropy (CE) loss component during the training of the {\tt CircuitDistilled-3B} model for entity tracking task, comparing the behavioral baseline ({\tt CE Only}) against our mechanistic method ({\tt CE + Align CKA}). The plot illustrates two key findings. 
First, the model trained with the {\tt CE + Align CKA} objective converges faster. The composite loss provides a richer, more structured gradient signal that the CE loss alone, which accelerates the learning process. For example, at the 1500-step mark the CE loss for mechanistic run was nearly half that of the behavioral baseline. Second, the mechanistic approach also leads to a lower final CE loss value by the end of the training run.

We hypothesize that this improved efficiency stems from the nature of the CKA loss term. While the CE loss provides a sparse signal based only on the final output token, the CKA loss offers a denser gradient that guides the internal representations of the student's circuit to align with the teacher's. This forces the student model to not just find the correct answer, but to adopt a proven, effective internal algorithm for doing so, thereby regularizing the learning process and leading to a more efficient and effective convergence. 

\subsection{Details for Sections \ref{sec:layer_vs_circuit} and \ref{sec:mech_transfer}}
\label{appx:mech_details}

\paragraph{Layer CKA} Each student layer $\ell$ that contains at least one circuit head (13 of 16 layers for {\tt Llama3-1B} on entity tracking) is paired with the teacher layer at the same relative depth, $\ell_t = \mathrm{round}\big(\ell \cdot (L_t - 1)/(L_s - 1)\big)$, where $L_s$ and $L_t$ are the numbers of student and teacher layers. The {\tt attn} variant aligns the concatenated head outputs at each layer (the input to $W_O$); the {\tt resid} variant aligns the residual stream at the layer output. We use the same $\lambda$ and update the same student parameters as in circuit distillation, so the two methods differ only in the alignment target.

\paragraph{KD} All student parameters are trained to match the teacher's temperature-scaled output distribution over response tokens,
\[ \mathcal{L}_{\text{KD}} = T^2 \cdot \mathrm{KL}\big(\mathrm{softmax}(z_t/T) \,\|\, \mathrm{softmax}(z_s/T)\big), \]
where $z_t$ and $z_s$ are the teacher and student logits and $T = 2$.

\paragraph{CKA analysis} We use 500 held-out entity tracking prompts, disjoint from the distillation data, and take per-head activations at the final token. The 34 mapped pairs are those produced by the ablation-based mapping (Section \ref{sec:distill-obj}). The 100 control pairs are sampled uniformly at random from the (student circuit head, teacher circuit head) pairs that are \emph{not} mapped to each other.

\paragraph{Attention analysis} We use 200 held-out prompts and locate the queried object by matching the context clause ``the $\langle$object$\rangle$ is in Box $\langle$label$\rangle$''. We select the three mapped pairs whose student and teacher heads are both among the 20 highest-scoring Value Fetcher heads in their respective models, ranked by the teacher head's patching score, and report the attention weight from the final position to the object token.

\paragraph{Path patching} We follow the procedure of Appendix \ref{appx:circuit_id} with the same data and settings (300 examples, 7 boxes), running only the first (Value Fetcher) stage, which takes about 35 minutes per {\tt Llama3-1B} model on one A100. \emph{VF overlap@20} is the fraction of the 20 highest-scoring heads that belong to the student circuit; Spearman $\rho$ is computed over the 34 mapped pairs.

\begin{figure}[h]
\centering
  \includegraphics[scale=0.9]{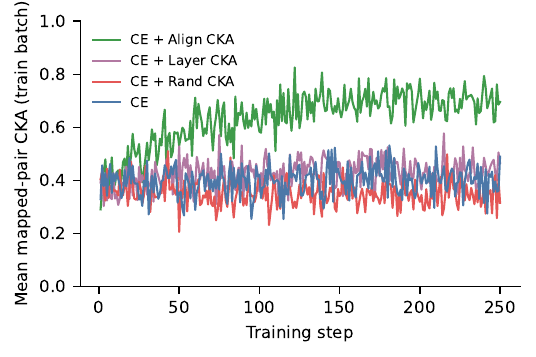}
  \caption{Mean CKA between mapped student and teacher circuit heads on training batches, over training of the {\tt Llama3-1B} entity tracking student.}
  \label{fig:appx_cka_curves}
\end{figure}

\paragraph{CKA during training} Figure \ref{fig:appx_cka_curves} tracks the mean CKA of the mapped head pairs on training batches. It rises steadily only under {\tt CE + Align CKA}; {\tt CE}, {\tt CE + Rand CKA}, and {\tt CE + Layer CKA} change it only slightly from its initial value.

\end{document}